\documentclass[11pt,a4paper]{article}
\usepackage[hyperref]{emnlp2020}
\usepackage{times}
\usepackage{latexsym}

\usepackage{tikz}
\usepackage{pgfplots}
\usepackage{subcaption}
\usepackage{lipsum}
\usepackage[draft]{todonotes}
\usepackage{amsmath}
\usepackage{amsfonts}
\usepackage{graphicx}
\usepackage{amssymb}
\usepackage{multirow, tabu}

\usepackage{microtype}

\usepackage{multirow}

\aclfinalcopy 

\title{Conversational Semantic Parsing for Dialog State Tracking}

\author{{Jianpeng Cheng\quad Devang Agrawal\quad Héctor Martínez Alonso\quad Shruti Bhargava\quad Joris Driesen} \\ 
        {\bf Federico Flego\quad Shaona Ghosh \quad Dain Kaplan\quad Dimitri Kartsaklis\quad Lin Li\quad Dhivya Piraviperumal } \\
        {\bf Jason D Williams\quad Hong Yu\quad Diarmuid Ó Séaghdha\quad Anders Johannsen} \\
     Apple \\
\texttt{\{jianpeng.cheng,hmartinezalonso,shruti\_bhargava,joris\_driesen,}\\
\texttt{fflego,shaona\_ghosh,dain\_kaplan,dkartsaklis,lli9,dhivyaprp,jason\_williams4,}\\
\texttt{hong\_yu,doseaghdha,ajohannsen\}@apple.com}
}

\date{}

\begin{document}
\maketitle
\begin{abstract}
We consider a new perspective on dialog state tracking (DST), the task of estimating a user's goal through the course of a dialog. By formulating DST as a semantic parsing task over hierarchical representations, we can incorporate semantic compositionality, cross-domain knowledge sharing and co-reference. 
We present TreeDST, a dataset of 27k conversations annotated with tree-structured dialog states and system acts.\footnote{The dataset is available at \url{https://github.com/ apple/ml-tree-dst}.}
We describe an encoder-decoder framework for DST with hierarchical representations, which leads to ~20\% improvement over state-of-the-art DST approaches that operate on a flat meaning space of slot-value pairs.
\end{abstract}

\section{Introduction}

Task-based dialog systems, for example digital personal assistants, provide a linguistic user interface for all kinds of applications: from searching a database, booking a hotel, checking the weather to sending a text message. In order to understand the user, the system must be able to both parse the meaning of an utterance and relate it to the context of the conversation so far. While lacking the richness of a conversation between two humans, the dynamics of human-machine interaction can still be complex: the user may change their mind, correct a misunderstanding or refer back to previously-mentioned information.

Language understanding for task-based dialog is often termed ``dialog state tracking'' (DST) \cite{williams2016dialog}, the mental model being that the intent of the user is a partially-observed state that must be re-estimated at every turn given new information. The dialog state is typically modelled as a set of independent slots, and a standard DST system will maintain a distribution over values for each slot. In contrast, language understanding for other NLP applications is often formulated as semantic parsing, which is the task of converting a single-turn utterance to a  graph-structured meaning representation. Such meaning representations include logical forms, database queries and other programming languages.




These two perspectives on language understanding---DST and semantic parsing---have complementary strengths and weaknesses. DST targets a fuller range of conversational dynamics but typically uses a simple and limiting meaning representation. Semantic parsing embraces a \textit{compositional} view of meaning. By basing meaning on a space of combinable, reusable parts, compositionality can make the NLU problem more tractable (repeated concepts must only be learned once) and more complex (it becomes possible to represent nested intents). At the same time, most semantic parsing research treats a sentence as an isolated observation, detached from conversational context.

\begin{table}[t!]
\footnotesize
\centering
\begin{tabular}{ p{0.03\textwidth} | p{0.4\textwidth} } 
 \hline
 \textbf{Turn} & \multicolumn{1}{c}{\textbf{Utterance and Annotation}}  \\ \hline
\multirow{7}{*}{1}  & \textcolor{red}{\textit{Hi can you book me a flight to Paris please.}} \newline \scriptsize{\texttt{user.flight.book.object.equals \newline \null\quad.destination.equals.location.equals.Paris}} \\ 
   & \textcolor{blue}{\textit{Sure, when and where will you depart?}} \newline  \scriptsize{\texttt{system.prompt.flight.book.object.equals \newline \null\quad.source \newline \null\quad.departureDateTime}}  \\ \hline
\multirow{13}{*}{2}  & \textcolor{red}{\textit{Tomorrow from London}} \newline \scriptsize{\texttt{user.flight.book.object.equals \newline \null\quad.destination.equals.location.equals.Paris \newline \null\quad.source.equals.location.equals.London \newline \null\quad.departureDateTime.equals\newline \null\quad\quad.date.equals.definedValue.equals.Tomorrow}} \\ 
   & \textcolor{blue}{\textit{I found 5 flights for you. The earliest one departs at 10 AM with a cost of \pounds105. Would you like it?}} \newline \scriptsize{\texttt{system\newline \null\quad.prompt.flight.book.object.equals\newline \null\quad\quad.departureDateTime.equals.time \newline\null\quad.inform.flight.find\newline\null\quad\quad.count.equals.5\newline\null\quad\quad.object.equals\newline\null\quad\quad\quad.departureDateTime.equals.time.equals\newline\null\quad\quad\quad\quad.hour.equals.10\newline\null\quad\quad\quad\quad.meridiem.equals.AM \newline\null\quad\quad\quad.price.equals.105}} \\ \hline
\multirow{6}{*}{3}  & \textcolor{red}{\textit{Do I have any calendar event on that day?}} \newline \scriptsize{\texttt{user.calendarEvent.checkExistence.object.equals\newline \null\quad.dateTimeRange.equals\newline \null\quad\quad.date.equals.definedValue.equals.Tomorrow}} \\ 
   & \textcolor{blue}{\textit{No you don't have any event.}} \newline \scriptsize{\texttt{system.inform.calendarEvent.find.notExisted}} \\ \hline
\multirow{11}{*}{4}  & \textcolor{red}{\textit{I would book the 10 AM flight for me please}} \newline  \scriptsize{\texttt{user.flight.book.object.equals \newline \null\quad.destination.equals.location.equals.Paris \newline \null\quad.source.equals.location.equals.London \newline \null\quad.departureDateTime.equals\newline \null\quad\quad.date.equals.definedValue.equals.Tomorrow\newline \null\quad\quad.time.equals\newline\null\quad\quad\quad.hour.equals.10\newline\null\quad\quad\quad.meridiem.equals.AM}}  \\ 
   & \textcolor{blue}{\textit{Here is your booking information. Please confirm.}} \newline \scriptsize{\texttt{system.offer.flight.book}} \\ \hline

\multirow{7}{*}{5} & \textcolor{red}{\textit{Direction to my next meeting.}} \newline
\scriptsize{\texttt{user.navigation.find.object.equals\newline \null\quad.destination.equals.reference\newline \null\quad\quad.calendarEvent.object.equals\newline \null\quad\quad\quad.listOffset.equals.1}} \\
& \textcolor{blue}{\textit{Here is the direction to your meeting at the GlassHouse.}} \newline \scriptsize{\texttt{system\newline \null\quad.navigation.inform.find\newline \null\quad.calendarEvent.inform.find.object.equals\newline \null\quad\quad.location.equals.GlassHouse}} \\
\hline
\end{tabular}
\caption{An example conversation in TreeDST with annotations. User and system utterances are marked in red and blue respectively. We use \textit{dot} to represent tree edges and an. increased indentation level to reveal multiple children attached to the same parent node. A side-by-side comparison between a dotted tree and its full rawing can be found in the Appendix.}
\label{examples}
\end{table}


This work unifies the two perspectives by reformulating DST as \textit{conversational semantic parsing}. As in DST, the task is to track a user's goal as it accumulates over the course of a conversation. The goal is represented using a structured formalism like those used in semantic parsing. Specifically, we adopt a hierarchical representation which captures domains, verbs, operators and slots within a rooted graph grounded to an ontology. The structured dialog state is capable of tracking nested intents and representing compositions in a single graph (Turn 5 Table \ref{examples}). The formalism also naturally supports cross-domain slot sharing and cross-turn co-reference through incorporating the shared slots or the references as sub-graphs in the representation (Turn 3 Table \ref{examples}).

Using a reverse annotation approach inspired by \citet{shah2018building} and \citet{rastogi2019towards}, we have collected a large dataset of task-oriented dialogs annotated with hierarchical meaning representations. Each dialog was generated through a two-step process. First, a generative dialog simulator produces a meaningful conversational flow and a template-based utterance for each turn in the conversation. Then the utterances are paraphrased by human annotators to render more realistic and natural conversations. 
The resulting dataset, which we call TreeDST, covers 27k conversations across 10 domains. 
Conversations in TreeDST are non-linear: they contain glitches which represent system failures and uncooperative user behaviors such as under- and over-specifying slot information.
There are also use cases not addressed in existing slot-filling datasets, including compositional intents and multi-intent utterances.

The second contribution of this work is a conversational semantic parser that tackles DST as a graph generation problem. At each turn, the model encodes the current user utterance and representations of dialog history, based upon which the decoder generates the updated dialog state with a mixture of generation and copy mechanism.
In order to track practical conversations with intent switching and resumption \cite{lee2016task,el2017frames},  we adopt a stack \cite{rudnicky1999agenda} to represent dialog history, and a parent-pointer decoder to speed up decoding.
We conducted controlled experiments to factor out the impact of hierarchical representations from model architectures. 
Overall our approach leads to 20\% improvement over state-of-the-art DST approaches that operate on a flat meaning space.


\section{Related Work}

\paragraph{Modeling}
Traditional DST models apply discriminative classifiers over the space of slot-value combinations \cite{crook2010representing,henderson2014word,williams2016dialog}. These models require feature extraction from user utterances based on manually constructed semantic dictionaries, making them
vulnerable to language variations. Neural classification models \cite{mrkvsic2017neural,mrkvsic2018fully} alleviate the problem by learning distributed representations of user utterances. However, they still lack scalability to large unbounded output space \cite{xu2018end,lee2019sumbt} and structured representations.
To address the limitations, some recent work treats slot filling as a sequence generation task  \cite{ren2019scalable,wu2019transferable}.

On the other hand, single-turn semantic parsers have long used structured meaning representations 
to address compositionality \cite{liang2013learning,banarescu2013abstract,kollar2018alexa,yu2018spider,gupta2018semantic}.
Solutions range from chart-based constituency parsers \cite{berant-etal-2013-semantic} to more recent neural sequence-to-sequence models \cite{jia2016data, dong2016language}. The general challenge of scaling semantic parsing to DST is that dialog state, as an accumulation of conversation history, requires expensive context-dependent annotation. It is also unclear how utterance semantics can be aggregated and maintained in a structured way. In this work we provide a solution to unify DST with semantic parsing.


\paragraph{Data Collection} The most straightforward approach to building datasets for task-oriented dialog is to directly annotate human-system conversations \cite{williams2016dialog}.
A limitation is that the approach requires a working system at hand, which causes a classic chicken-and-egg problem for improving user experience.
The issue can be avoided with Wizard-of-Oz (WoZ) experiments to collect human-human conversations \cite{el2017frames,budzianowski2018multiwoz,peskov2019multi,byrne2019taskmaster,radlinski2019coached}. However, 
while dialog state annotation remains challenging and costly in WoZ, the resulting distribution could be different from that of 
human-machine conversations \cite{budzianowski2018multiwoz}.
One approach that avoids direct meaning annotation is to use a dialog simulator \cite{schatzmann2007agenda,li2016user}.
Recently, \citet{shah2018building} and \citet{rastogi2019towards} generate synthetic conversations which are subsequently paraphrased by crowdsourcing. This approach has been proven to provide a better coverage while reducing the error and cost of dialog state annotation \cite{rastogi2019towards}. We adopt a similar approach in our work, but focusing on a structured meaning space.


\section{Setup}
\subsection{Problem Statement}
We use the following notions throughout the paper.
A conversation $X$ has representation $Y$ grounded to an ontology $\mathcal{K}$:
at turn $t$, every user utterance $x^u_t$ is annotated with a \textit{user dialog state} $y^u_t$, which represents an accumulated user goal up to the time step $t$. 
Meanwhile, every system utterance $x^s_t$ is annotated with a \textit{system dialog act} $y^s_t$, which represents the system action in response to $y^u_t$.
Both $y^u_t$ and $y^s_t$ adopt the same structured semantic formalism to encourage knowledge sharing between the user and the system.
From the perspective of the system, $y^s_t$ is observed (the system knows what it has just said) and $y^u_t$ must be inferred from the user's utterance.
For the continuation of an existing goal, the old dialog state will keep being updated; 
however, when the user proposes a completely new goal during the conversation, a new dialog state will overwrite the old one. A stack is used to store non-accumulable dialog states in the entire conversation history (in both data simulation and dialog state tracking). 

There are two missions of this work: 1) building a conversational dataset with structured annotations that can effectively represent the joint distribution $P(X, Y)$;
and 2) building a dialog state tracker which estimates the conditional distribution of every dialog state given the current user input and dialog history $P(y^u_t |x^u_t, X_{<t}, Y_{<t})$.

\subsection{Representation}
We adopt a hierarchical and recursive semantic representation for user dialog states and system dialog acts.
Every meaning is rooted at either a \texttt{user} or \texttt{system} node to distinguish between the two classes.
Non-terminals of the representation include \texttt{domains}, user \texttt{verbs}, system \texttt{actions}, \texttt{slots}, and \texttt{operators}.
A \texttt{domain} is a  group of activities such as creation and deletion of calendar events.
A user \texttt{verb} represents the predicate of the user intent sharable across domains, such as \texttt{create}, \texttt{delete}, \texttt{update}, and \texttt{find}.
A system \texttt{action} represents a type of system dialog act in response to a user intent. For example, the system could \texttt{prompt} for a slot value; \texttt{inform} the user about the information they asked for; and \texttt{confirm} if an intent or slot is interpreted correctly.
Nested properties (e.g., time range) are represented as a hierarchy of \texttt{slot}-\texttt{operator}-argument triples,
where the argument 
can be either a sub-slot, a terminal \texttt{value} node or a special \texttt{reference} node.
The \texttt{value} node accepts either a categorical label (e.g., day of week) or an open value (e.g., content of text message).
The \texttt{reference} node allows a whole intent to be attached as a slot value, enabling the construction of cross-domain use cases (e.g., Turn 5 of Table \ref{examples}). Meanwhile, co-reference to single slots is directly achieved by subtree copying, as shown in Turn 3.
Finally, conjunction is implicitly supported by allowing a set of arguments to be attached to the same slot (see Appendix for details).
Overall, the representation presented above focuses on ungrounded utterance semantics to decouple understanding from execution.
By incorporating domain-specific logic, the representation can be mapped to executable programs.

\section{Data Elicitation}
\subsection{Overview}
We sidestep the need for collecting real human-system conversations and annotating them with complex semantics by adopting a reverse data elicitation approach \cite{shah2018building,rastogi2019towards}. We model the generative process $P(X, Y) = P(Y)P(X|Y)$,
where representations of conversation flows ($Y$) are firstly rendered by a dialog simulator, and then realised into natural dialog ($X$) by annotators.
Two central aspects which directly impact the quality of the resulting data are: (1) the dialog simulator which controls the coverage and naturalness of the conversations; (2) the annotation instructions and quality control mechanism that ensures $X$ and $Y$ are semantically aligned after annotation.

\subsection{Simulator}
The most common approach of simulating a conversation flow is agenda-based \cite{schatzmann2007agenda,li2016user,shah2018building,rastogi2019towards}. 
At the beginning of this approach, a new goal is defined in the form of slot-value pairs describing user’s requests and constraints; and an agenda is constructed by decomposing the user goal into a sequence of user actions. Although the approach ensures the user behaves in a goal-oriented manner, it constrains the output space with \textit{pre-defined} agendas, which is hard to craft for complex user goals \cite{shi2019build}.

Arguably, a more natural solution to dialog simulation for complex output space is a fully generative method.
It complies with the behavior that a real user may only have an initial goal at the start of conversation, while the final dialog state cannot be foreseen in advance. 
The whole conversation can be defined generatively as follows:
\begin{equation}
P(Y) = P(y^u_0)  \sum_{t=0}^n P(y^s_t | y^u_t) \sum_{t=1}^n P(y^u_t | y^s_{<t}, y^u_{<t})
\end{equation}
where $Y$ is the conversation flow, $y^u_t$ is the user dialog state at turn $t$ and $y^s_t$ the system dialog act.
The decomposed probability of $P(Y)$ captures the functional space of dialog state transitions 
with three components: 1) a module generating the initial user goal $P(y^u_0)$,
2) a module generating system act $P(y^s_t | y^u_t)$, and 3) a module for user state update based on the dialog history  $P(y^u_t | y^s_{<t}, y^{u}_{<t})$. 
The conversation terminates at time step $n$ which must be a finishing state (system success or failure). 

\paragraph{Initial intent module $P(y^u_0)$} The dialog state $y^u_0$ representing the initial user goal is generated with a probabilistic tree substitution grammar \cite[PTSG]{cohn2010inducing} based on our semantic formalism. 
Non-terminal symbols in the PTSG can rewrite entire tree fragments to encode non-local context such as hierarchical and co-occurring slot combinations.
As explained in the example below, the algorithm generates dialog states from top down by recursively substituting non-terminals (marked in green) with subtrees.

\definecolor{celadon}{rgb}{0.67, 0.88, 0.69}
\renewenvironment{quote}
  {\small\list{}{\rightmargin=.1cm \leftmargin=.1cm}%
   \item\relax}
  {\endlist}
\begin{quote}
\small
\begin{tabular}{ p{0.13\textwidth} p{0.5\textwidth}}
\texttt{\$createEvent} & $\rightarrow$\,\texttt{user.calendarEvent.create\newline\null\quad\quad\quad.\colorbox{celadon}{\$newEvent}}
\end{tabular}

\begin{tabular}{ p{0.093\textwidth} p{0.3\textwidth}}
\texttt{\$newEvent} & $\rightarrow$\,\texttt{object.equals\newline\null\quad\quad\quad.\colorbox{celadon}{\$attendees}\newline\null\quad\quad\quad.\colorbox{celadon}{\$location}}
\end{tabular}

\end{quote}
This example generates a calendar event creation intent that contains two slots of attendees and location. In a statistical approach, the sampling probability of each production rule could be learned from dialog data.
For the purpose of bootstrapping a new model, any prior distribution can be applied.

\paragraph{Response module $P(y^s_t | y^u_t)$} 
At turn $t$, the system act $y^s_t$ is generated based on the current user dialog state $y^u_t$ and domain-specific logic.
We define an interface for the conditional generation task with a probabilistic tree transformation grammar,
which captures how an output tree ($y^s_t$) is generated from the input ($y^u_t$) through a mixture of generation and a copy mechanism.
As shown in the example below, every production rule in the grammar is in the form $A \rightarrow B$, where $A$ is an input \textit{pattern} that could contain both observed nodes and unobserved nodes (marked in red),
and $B$ is an output \textit{pattern} to generate.

\begin{quote}
\small
\begin{tabular}{p{.5\textwidth}}
\center
\texttt{user.calendarEvent.create \newline\null\quad .object.equals \newline\null\quad\quad .\colorbox{pink}{-dateTimeRange}}\\ 
$\big\downarrow$  \\
\texttt{system.prompt.calendarEvent\newline\null\quad\quad.create.object.equals \newline\null\quad\quad\quad.dateTimeRange}
\end{tabular}
\end{quote}

Given a user dialog state, the simulator \textit{looks up} production rules which result in a match of pattern $A$, and then derives a system act based on the pattern $B$.
Like in PTSG, probabilities of transformations can be either learned from data or specified a priori.

\paragraph{State update module $P(y^u_t | y^s_{<t}, y^{u}_{<t})$}
The generation of the updated dialog state is dependent on the dialog history.
While the full space of $y^u_t$ is unbounded, we focus on simulating three common types of updates, where a user introduces a new goal, continues with the previous goal $y^u_{t-1}$, or resumes an earlier unfinished goal (see Turn 2-4 of Table \ref{examples} for examples, respectively).
To model dialog history, we introduce an \textit{empty} stack when the conversation starts. A dialog state and the rendered system act are pushed onto the stack upon generation, dynamically updated during the conversation, and popped from the stack upon task completion. Therefore, the top of the stack always represents the most recent uncompleted task $y^{top, u}_{t-1}$ and the corresponding system act $y^{top, s}_{t-1}$. 

We consider the top elements of the stack as the effective dialog history and use it to generate the next dialog state $y^u_t$.
The generation interface is modeled with a similar tree transformation grammar, but every production rule has two inputs in the form $A, B \rightarrow C$: 

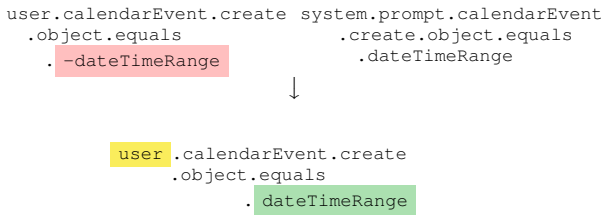
\definecolor{corn}{rgb}{0.98, 0.93, 0.36}
\renewenvironment{quote}
  {\small\list{}{\rightmargin=.1cm \leftmargin=.1cm}%
   \item\relax}
  {\endlist}
\begin{quote}
\scriptsize
\begin{tabular}{ p{0.215\textwidth} p{0.4\textwidth}}
\texttt{user.calendarEvent.create \newline\null\quad .object.equals \newline\null\quad\quad .\colorbox{pink}{-dateTimeRange}} &  \texttt{system.prompt.calendarEvent\newline\null\quad\quad.create.object.equals \newline\null\quad\quad\quad.dateTimeRange}\\
\multicolumn{2}{l}{\quad\quad\quad\quad\quad\quad\quad\quad\quad\quad\quad\quad\quad\quad\quad$\big\downarrow$}
\end{tabular}
\begin{tabular}{p{.5\textwidth}}
\center
\texttt{\colorbox{corn}{user}.calendarEvent.create \newline\null\quad .object.equals \newline\null\quad\quad .\colorbox{celadon}{dateTimeRange}}
\end{tabular}
\end{quote}
where $A$ specifies a matching pattern of the user goal $y^{top, u}_{t-1}$, $B$ is a matching pattern of the system act $y^{top, s}_{t-1}$, and $C$ is an output pattern that represents how the updated dialog state is obtained through a mixture of grammar expansion (marked in green) and copy mechanism from either of the two sources (marked in yellow).

\subsection{Annotation}
Following \citet{shah2018building,rastogi2019towards}, every grammar production in the simulator is paired with a template whose slots are synchronously expanded. As a result, each dialog state or system act is associated with a template utterance.
The purpose is to offer minimum understandability to each conversation flow, based on which annotators will generate natural conversations.

\paragraph{Instructions} Annotators generate a conversation based on the given templated utterances.
The task proceeds turn by turn. For each turn, we instruct annotators to convey exactly the same intents and slots in each user or system utterance, in order to make sure the obtained utterance agrees with the programmatically generated semantic annotation. 
The set of open values (specially marked in brackets, such as event titles and text messages) must be preserved too.
Besides the above restrictions, we give annotators the freedom to generate an utterance with paraphrasing, compression and expansion in the given dialog context, to make conversations as natural as possible. While system utterances are guided to be expressed in a professional tone, we encourage annotators to introduce adequate syntactic variations and chit-chats in user utterances as long as they do not change the intent to be delivered.

\paragraph{Quality control} We enforce two quality control mechanisms before and after the dialog rendering task.
Before the task, we ask annotators to provide a binary label to each conversation flow.  The label indicates if the conversation contains any non-realistic interactions. We can therefore filter out low-quality data outputted by the simulator. After the task, we ask a different batch of annotators to evaluate if each human-generated utterance preserves the meaning of the templated utterance; any conversation that fails this check is removed.


\subsection{Statistics}
The resulting TreeDST dataset consists of 10 domains: \textit{flight}, \textit{train}, \textit{hotel}, \textit{restaurant}, \textit{taxi}, \textit{calendar}, \textit{message}, \textit{reminder}, \textit{navigation}, and \textit{phone},  exhibiting nested properties for \textit{people}, \textit{time}, and \textit{location} that are shared across domains.
Table \ref{comparison} shows a comparison of TreeDST with the following pertinent datasets, \textbf{DSTC2} \cite{henderson-etal-2014-second},
\textbf{WOZ2.0} \cite{Wen2017A},
\textbf{FRAMES} \cite{el2017frames},
\textbf{M2M} \cite{shah2018building},
\textbf{MultiWOZ} \cite{budzianowski2018multiwoz} and
\textbf{SGD} \cite{rastogi2019towards}.
Similar to our work, both \textbf{M2M} and \textbf{SGD} use a simulator to generate conversation flows; and both \textbf{MultiWOZ} and \textbf{SGD} contain multi-domain conversations. The difference is that all the previous work represents dialog states as flat slot-value pairs, which are not able to capture complex relations such as compositional intents.

\begin{table*}[t!]
\centering
\small
\begin{tabular}{ |l|c|c|c|c|c|c|c| } 
 \hline
  & \textbf{DSTC2} & \textbf{WOZ2.0} & \textbf{FRAMES} & \textbf{M2M} & \textbf{MultiWOZ} & \textbf{SGD} & \textbf{TreeDST} \\ \hline
  Representation & \multicolumn{6}{c|}{Flat} & Hierarchical \\ \hline
 \#Dialogs & 1,612 & 600 & 1,369 & 1,500 & 8,438 & 16,142 & 27,280  \\ \hline
 Total \#turns & 23,354 & 4,472 & 19,986 & 14,796 & 113,556 & 329,964 & 167,507 \\ \hline
 Avg. \#turns/dialog & 14.5 & 7.45 & 14.60 & 9.86 & 13.46 & 20.44 & 7.14 \\ \hline
 Avg. \#tokens/utterance & 8.54 & 11.24 & 11.24 & 8.24 & 13.13 & 9.75 &  7.59\\ \hline
 \#slots & 8 & 4 & 61 & 13 & 24 & 214 & 287 \\ \hline
 \#values & 212 & 99 & 3,871 & 138 & 4,510 & 14,139 & 20,612 \\ \hline
 \#multi-domain dialog & - & - & - & - & 7,032 & 16,142 & 14,999  \\ \hline
 \#compositional utterance & - & - & - & - & - & - & 10,133 \\ \hline
 \#cross-turn co-reference & - & - & - & - & - & - & 9,609 \\ 
 \hline
\end{tabular}
\caption{Comparison of TreeDST with pertinent datasets for task-oriented dialogue.}
\label{comparison}
\end{table*}

\section{Dialog State Tracking}
The objective in the conversational semantic parsing task is to predict the updated dialog state at each turn given the current user input and dialog history $P(y^u_t |x^u_t, X_{<t}, Y_{<t})$.
We tackle the problem with an encoder-decoder model: at turn $t$, the model encodes the current user utterance $x^u_t$ and dialog history, conditioned on which the decoder predicts the target dialog state $y^u_t$. We call this model the Tree Encoder-Decoder, or \textsc{Ted}.

\paragraph{Dialog history} 
When a dialog session involves task switching, there will be multiple, non-accumulable dialog states in the conversation history.
Since it is expensive to encode the entire history $X_{<t}, Y_{<t}$ whose size grows with the conversation,
we compute a fixed-size history representation derived from the previous conversation flow ($Y_{<t}$).
Specifically, we reuse the notation of a stack to store past dialog states, and the top of the stack $y^{top, u}_{t-1}$ tracks the most recent uncompleted task.
The dialog history is then represented with the last dialog state $y^u_{t-1}$, the dialog state on top of the stack $y^{top, u}_{t-1}$, and the last system dialog act $y^s_{t-1}$. We merge the two dialog states $y^u_{t-1}$ and $y^{top, u}_{t-1}$ into a single tree $Y^u_{t-1}$ for featurization.

\begin{figure*}[t]
  \centering
  \includegraphics[width=\columnwidth*2]{fig/encdec.jpg}
  \caption{An overview of the \textsc{Ted} encoder-decoder architecture.}
  \label{DST}
\end{figure*}

\paragraph{Encoding} We adopt three encoders for utterance $x^u_t$, system act $y^s_{t-1}$ and dialog state $Y^u_{t-1}$ respectively. 
For the user utterance $x^u_t$, a bidirectional LSTM encoder is used to convert the word sequence
into an embedding list $\mathbf{H_x} = [\mathbf{h^x_1}, \mathbf{h^x_2}, \cdots, \mathbf{h^x_n}]$, where $n$ is the length of the word sequence. For both the previous system act $y^s_{t-1}$ and user state $Y^u_{t-1}$, we linearize them into strings through depth-first traversal (see Figure \ref{DST}).
Then the linearized $y^s_{t-1}$ and $Y^u_{t-1}$ are encoded with two separate bidirectional LSTMs.
The outputs are two embedding lists: $\mathbf{H_s} =  [\mathbf{h^s_1}, \mathbf{h^s_2}, \cdots, \mathbf{h^s_m}]$ where $m$ is the length of the  linearized system act sequence, and $\mathbf{H_u} =  [\mathbf{h^u_1}, \mathbf{h^u_2}, \cdots, \mathbf{h^u_l}]$ where $l$ is the length of the linearized dialog state sequence.
The final outputs of encoding are $\mathbf{H_x}$, $\mathbf{H_s}$ and $\mathbf{H_u}$.

\paragraph{Decoding}
After encoding, the next dialog state $y^u_t$ is generated with an LSTM decoder as a linearized string which captures the depth-first traversal of the target graph (see Figure \ref{DST}).

At decoding step $i$, the decoder feeds the embedding of the previously generated token $y^u_{t, i-1}$ and updates the decoder LSTM state to $\mathbf{g_i}$:
\begin{equation}
    \mathbf{g_i} = \text{LSTM} (\mathbf{g_{i-1}}, \mathbf{y^u_{t, i-1}})
\end{equation}
An attention vector is computed between the state $\mathbf{g_i}$ and each of the three encoder outputs $\mathbf{H^x}$, $\mathbf{H^s}$ and $\mathbf{H^u}$.
For each of the encoder memory $\mathbf{H}$, the computation is defined as follows:
\begin{equation}
\begin{split}
    \mathbf{a_{i,j}} &= \textit{attn}(\mathbf{g_i}, \mathbf{H})\\
    \mathbf{w_{i,j}} &= \textit{softmax}(\mathbf{a_{i,j}}) \\
    \mathbf{\bar{h}_i} &= \sum_{j=1}^{n} \mathbf{w_{i,j}} \mathbf{h_j}
\end{split}
\end{equation}
\label{attention}
where \textit{attn} represents the feed-forward attention defined in \citet{bahdanau2015neural} and the \textit{softmax} is taken over index $j$.
 By applying the attention mechanism to all three sources, we get three attention vectors $\mathbf{\bar{h}^x_i}$, $\mathbf{\bar{h}^s_i}$, and $\mathbf{\bar{h}^u_i}$. 
The vectors are concatenated together with the state $\mathbf{g_i}$ to form a feature vector $\mathbf{f_i}$, which is used to compute the probability of the next token though a mixture of generation and copy mechanism \cite{gu2016incorporating}:
\begin{equation}
    \begin{split}
        \lambda & = \sigma(\mathbf{W_i}  \mathbf{f_i} + \mathbf{b_i}) \\
        P_{gen} & = \textit{softmax}(\mathbf{W_v}  \mathbf{f_i} + \mathbf{b_v}) \\
        P_{copy} & = \textit{softmax}(\mathbf{a_i}, \mathbf{c_i}, \mathbf{e_i} )  \\
        P(y^u_{t, i}) & = \lambda P_{gen} + (1-\lambda)  P_{copy} 
    \end{split}
    \label{copy mechanism}
\end{equation}
where $\mathbf{W}$ and $\mathbf{b}$ are all model parameters. $\lambda$ is a soft gate controlling the proportion of generation and copy. $P_{gen}$ is computed with a softmax over the generation vocabulary.
$\mathbf{a}$, $\mathbf{c}$ and $\mathbf{e}$ denote attention logits computed for the three encoders. Since there are three input sources, we concatenate all logits and normalize them to compute the copy distribution $P_{copy}$. The model is optimised on the log-likelihood of output distribution $P(y^u_{t, i})$.
An overview of the model is shown in Figure  \ref{DST}.

\subsection{Parent Pointer: a Faster Graph Decoder}
One observation about the standard decoder is that it has to predict long strings with closing brackets to represent a tree structure in the linearization. Therefore the total number of decoding LSTM recursions is the number of tree nodes plus the number of non-terminals. We propose a modified \textit{parent pointer} (PP) decoder which reduces the number of autoregressions to the number of tree nodes. This optimisation is not applicable only to our DST model, but to any decoder that treats tree decoding as sequence prediction in the spirit of \citet{VinyalsEtAl:2015}.

The central idea of the PP decoder is that at each decoding step, two predictions will be made: one generates the next tree node, and the other selects its parent from the existing tree nodes. Eventually $y^u_t$ can be constructed from a list of tree nodes $n^u_t$ and a list of parent relations $r^u_t$. More specifically, at time step $i$, the decoder takes in the embeddings of the previous node $n^u_{t, i-1}$ and its parent $r^u_{t, i-1}$ to generate the hidden state $g_i$. 
\begin{equation}
    \mathbf{g_i} = \text{LSTM} (\mathbf{g_{i-1}}, \mathbf{n^u_{t, i-1}},  \mathbf{r^u_{t, i-1}})
\end{equation}
This state is then used as as the input for two prediction layers.
The first layer predicts the next node probability $P(n^u_{t, i})$ with Equation \ref{attention} to \ref{copy mechanism},
and the second layer selects the parent of the node by attending $\mathbf{g_i}$ to the previously generated nodes, which are represented with the decoder memory $\mathbf{G_{i-1}} = [\mathbf{g_1}, \cdots, \mathbf{g_{i-1}}]$:
\begin{equation}
\begin{split}
    \mathbf{f_{i,j}} &= \textit{attn}(\mathbf{g_i}, \mathbf{G_{i-1}}) \\
    P(r^u_{t, i}) &= \textit{softmax}(\mathbf{f_{i,j}}) 
\end{split}
\end{equation}
The model is optimised on the average negative log-likelihood of distributions $P(n^u_{t, i})$ and  $P(r^u_{t, i})$.

\section{Experiments}

\paragraph{Setup}
We split the TreeDST data into train (19,808), test (3,739) and development (3,733) sets. For evaluation, we measure turn-level dialog state exact match accuracy averaged over all turns in the test set. 
We evaluate the proposed model with its ``vanilla'' decoder (\textsc{Ted-Vanilla}) and its parent-pointer variant (\textsc{Ted-PP}). In both cases, the utterance encoder has 2 layers of 500 dimensions; the system act encoder and dialog state encoder have 2 layers of 200 dimensions; and the decoder has 2 LSTM layers of 500 dimensions. Dimensions of word and tree node embeddings are 200 and 50 respectively. Training uses a batch size of 50 and Adam optimizer \cite{kingma2014adam}. Validation is performed every 2 epochs and the training stops when the validation error does not decrease in four consecutive evaluations. The hyper-parameters were selected empirically based on an additional dataset that does not overlap with TreeDST.

\paragraph{Baselines}
In order to factor out the contribution of meaning representations from model changes in experiments, we additionally derive a version of our dataset where all meaning representations are flattened into slot-value pairs (details are described in the next paragraph). 
We then introduce a baseline \textsc{Ted-Flat} by training the same model (as \textsc{Ted-Vanilla}) on the flattened dataset.

We additionally introduce as baselines two state-of-the-art slot-filling DST models based on encoder-decoders: they include \textsc{comer} \cite{ren2019scalable} which encodes the previous system response transcription and the previous user dialog state and decodes slot values;
and \textsc{trade} \cite{wu2019transferable} which encodes all utterances in the history. Since both \textsc{Ted-Flat} and the two baselines are trained with flattened slot-value representations, we can compare various models in this setup.

\paragraph{TreeDST flattening}
To flatten TreeDST we collapse each path from the \texttt{domain} to leaf nodes into a single slot. \texttt{Verb} nodes in the path are excluded to avoid slot explosion. Take the following tree as an example:

\noindent \footnotesize{\texttt{\newline user.flight.book.object.equals \newline \null\quad.source.equals.location.equals.London \newline \null\quad.departureDateTime.equals\newline \null\quad\quad.date.equals.definedValue.equals\newline \null\quad\quad\quad.Tomorrow \newline \null\quad\quad.time.equals.hour.equals.5 \newline}}

\noindent \normalsize{Three slot-value pairs can be extracted:}

\noindent \footnotesize{\texttt{\newline
(flight+object+source+location, London) \newline
(flight+object+departureDateTime+date\newline +definedValue, Tomorrow) \newline
(flight+object+departureDateTime+time\newline+hour, 5) \newline}

\noindent \normalsize{The} operator \texttt{equals} is not shown in the slot names to make the names more concise.

\begin{table}[t!]
\centering
\begin{tabular}{ |c|c| } 
 \hline
 \textbf{Decoders} & \textbf{Accuracy}  \\ \hline
 \textsc{Ted-Vanilla}  & 0.622  \\ \hline
 \textsc{Ted-PP} & 0.622  \\ \hline \hline
 \textsc{Ted-Flat} & 0.535 \\ \hline
 \textsc{comer} \cite{ren2019scalable} & 0.509 \\ \hline
 \textsc{trade} \cite{wu2019transferable} & 0.513 \\ \hline
\end{tabular}
\caption{Results on the TreeDST test set}
\label{result}
\end{table}

\begin{table}[t!]
\centering
\begin{tabular}{ |l| c | } 
 \hline
 \textbf{Pattern} & \textbf{Accuracy}  \\ \hline
 All turns & 0.647 \\ \hline
 Turns with intent change & 0.552 \\ \hline
 Turns with compositional utterances  & 0.602 \\ \hline
 Turns with multi-intent utterances & 0.478   \\ \hline
\end{tabular}
\caption{Results on the TreeDST development set, broken down by dialog phenomena}
\label{ea1}
\end{table}

\begin{figure}[t!]
\begin{tikzpicture}
\begin{axis}[
    width=8cm,
    height=4cm,
    axis lines=left,
    axis x line=middle,
    ymin=0,
    ymax=1.0,
    xlabel=Turn ID,
    ylabel=Accuracy,
    ylabel near ticks,
    xtick=data,
    nodes near coords,
    every node near coord/.append style={font=\tiny},
    extra x ticks=1,
    ticklabel style={font=\small}]
\addplot[color=black, mark=*, thick, mark options={scale=0.5}, every node near coord/.append style={yshift=-12pt}] coordinates {
    (1, 0.844) (2, 0.792) (3, 0.771) (4, 0.713) (5, 0.671) (6, 0.509) (7, 0.368) (8, 0.286) };
\addplot[color=red, mark=*, thick, mark options={scale=0.5}] coordinates {
    (1, 0.842) (2, 0.818) (3, 0.800) (4, 0.788) (5, 0.810) (6, 0.638) (7, 0.722) (8, 0.681) };
\node[anchor=west,color=red, font=\small] at (550,95){oracle};
\end{axis}
\end{tikzpicture}
\caption{Validation exact match accuracy by turn ID}
\label{ea2}
\end{figure}
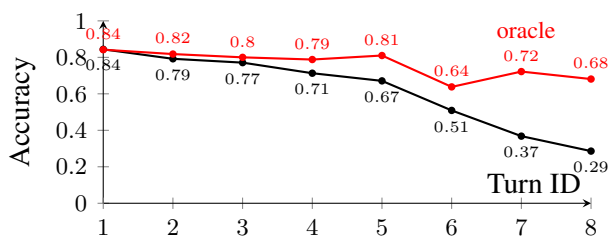

\paragraph{Results}
The results are shown in Table \ref{result}. 
Overall, both \textsc{Ted-Vanilla} and \textsc{Ted-PP} models bring 20\% relative improvement over existing slot-filling-based DST models.
By further factoring out the impact of representation and model differences, we see that representation plays a more important role:
the \textsc{Ted-Flat} variant, which differs only that it was trained on flattened parses, is clearly outperformed by \textsc{Ted-Vanilla} and \textsc{Ted-PP}.
We conclude that even if dialog states can be flattened into slot-value pairs, it is still more favorable to use a compact, hierarchical meaning representation.
The advantage of the representation is that it improves knowledge sharing across different domains (e.g., \texttt{message} and \texttt{phone}), verbs (e.g., \texttt{create} and \texttt{update}), and dialog participators (\texttt{user} and \texttt{system}). 
The second set of comparison is among different modeling approaches using the same flat meaning representation. \textsc{Ted-Flat} slightly
outperforms  \textsc{comer} and \textsc{trade}. The major difference is that our model encodes both past user and system representations; while the other models used past transcriptions. We believe the gain of encoding representations is that they are unambiguous; and the encoding helps knowledge sharing between the user and the system.

The vanilla and PP decoders achieve the same exact match accuracy. 
The average training time per epoch for PP is 1,794 seconds compared to 2,021 for vanilla, i.e. PP leads to 10\% reduction in decoding time without any effect on accuracy. We believe that the efficiency of PP can be further
optimized by parallellizing the two prediction layers of nodes and parents.

\paragraph{Analysis}
First, Table \ref{ea1} shows a breakdown of the \textsc{Ted-PP} model performance by dialog behavior on the development set. 
While the model does fairly well on compositional utterances, states with intent switching and multiple intents are harder to predict. We believe the reason is that the prediction of intent switching requires task reference resolution within the model; while multi-intent utterances tend to have more complex trees. 

Second, Figure \ref{ea2} shows the vanilla model results by turn index on the development set (black curve). This shows the impact of error propagation as the model predicts the target dialog state based on past representations. To better understand the issue, we compare to an oracle model which always uses the gold previous dialog state for encoding (red curve).

Vanilla model accuracy decreases with turn index, resulting in a gap with the oracle model. The error propagation problem can be alleviated by providing more complete dialog history to the encoder for error recovery \cite{henderson2014word}, which we consider as future work. 

Finally, we would like to point out a limitation of our approach in tracking dialog history with a stack-based memory. While the stack is capable of memorizing and returning to a previously unfinished task, there are patterns which cannot be represented such as switching between two ongoing tasks. We aim to explore a richer data structure for dialog history in the future.

\section{Conclusion}
This work reformulates dialog state tracking as a conversational semantic parsing task to overcome the limitations of slot filling.
Dialog states are represented as rooted relational graphs to encode compositionality, and encourage knowledge sharing across different domains, verbs, slot types and dialog participators.
We demonstrated how a dialog dataset with structured labels for both user and system utterances can be collected with the aid of a generative dialog simulator.
We then proposed a conversational semantic parser that does dialog state tracking through a stack-based memory, a mixture of generation and copy mechanism, and a parent-pointer decoder that speeds up tree prediction.
Experimental results show that our DST solution outperforms slot-filling-based trackers by a large margin.

\bibliography{emnlp2020}
\bibliographystyle{acl_natbib}

\end{document}